\documentclass[lettersize,conference]{ieeeconf}

\IEEEoverridecommandlockouts                              
                                                          
\usepackage{cite}
\usepackage{amsmath,amssymb,amsfonts}
\usepackage{algorithmic}
\usepackage{graphicx}
\usepackage{textcomp}
\usepackage{xcolor}
\def\BibTeX{{\rm B\kern-.05em{\sc i\kern-.025em b}\kern-.08em
    T\kern-.1667em\lower.7ex\hbox{E}\kern-.125emX}}
    
\usepackage{xspace}
\let\OldTexttrademark\texttrademark
\renewcommand{\texttrademark}{\OldTexttrademark\xspace }%

\usepackage [english]{babel}
\usepackage [autostyle, english = american]{csquotes}
\MakeOuterQuote{"}
    
\usepackage{hyperref}
\newcommand{\email}[1]{\href{mailto:#1}{#1}}

\usepackage{bm}

\usepackage{cleveref}
\usepackage{afterpage}
\usepackage{gensymb}
\usepackage{paralist}
\usepackage{color}

\newtheorem{definition}{\textbf{Definition}}

\newtheorem{theorem}{\rm\textbf{Theorem}}

\renewcommand{\baselinestretch}{0.915} 

\begin{document}


\title{Safe Control for Soft-Rigid Robots with Self-Contact \\using Control Barrier Functions}

\author{Zach J. Patterson$^{1}$, Wei Xiao$^{1}$, Emily Sologuren$^{1}$, and Daniela Rus$^{1}$
\thanks{$^{1}$ Computer Science and Artificial Intelligence Laboratory, MIT. \email{zpatt@mit.edu}, \email{weixy@mit.edu}, \email{rus@csail.mit.edu}}%
}

\maketitle

\begin{abstract}
Incorporating both flexible and rigid components in robot designs offers a unique solution to the limitations of traditional rigid robotics by enabling both compliance and strength. This paper explores the challenges and solutions for controlling soft-rigid hybrid robots, particularly addressing the issue of self-contact. Conventional control methods prioritize precise state tracking, inadvertently increasing the system's overall stiffness, which is not always desirable in interactions with the environment or within the robot itself. To address this, we investigate the application of Control Barrier Functions (CBFs) and High Order CBFs to manage self-contact scenarios in serially connected soft-rigid hybrid robots. Through an analysis based on Piecewise Constant Curvature (PCC) kinematics, we establish CBFs within a classical control framework for self-contact dynamics. Our methodology is rigorously evaluated in both simulation environments and physical hardware systems. The findings demonstrate that our proposed control strategy effectively regulates self-contact in soft-rigid hybrid robotic systems, marking a significant advancement in the field of robotics.
\end{abstract}

\section{Introduction}

As the soft robotics field continues its growth towards maturity, there is a nascent trend towards soft-rigid hybrid robot forms to allow both compliance for safe operation in uncertain environments and rigidity to allow load bearing capability 
 \cite{bernSimulation2022,zhuSoftRigid2023,coevoetPlanning2022,zhangGeometric2020}. Indeed, the majority of larger life forms (mammals, reptiles, birds, amphibians, and even fish) have some articulated rigid body structure that allows self-support under gravity. Such robots may expand the range of potential behaviors of robots, but they also may instantiate in new problems. In this work, we will look at a class of soft-rigid robots that frequently undergo rigid self contact and we will seek to control these systems to gracefully deal with self contact.

While the field of soft robotic control has experienced rapid growth over the past decade \cite{dellasantinaModelBasedControlSoft2023}, critical open questions remain. Thus far, most works have focused on precise state \cite{pattersonRobust2022} or end-effector tracking \cite{dellasantinaModelbased2020}, yet these tasks may ultimately be incidental to the goals of soft robots. This is because, as discussed in \cite{dellasantinaControlling2017}, there is a trade-off between feedback and stiffness, with more feedback increasing the effective stiffness of the system, eliminating the benefits of soft materials. Feedforward control does not suffer the same issues, but requires precise models that are fundamentally difficult for soft robots. Thus, while balanced feedforward plus feedback controllers like the PD+ controller have been shown to stabilize state trajectories for soft robots \cite{dellasantinaModelBasedControlSoft2023}, this comes at a cost of stiffening the robot's potential interactions with the environment (or with itself in self-contact) \cite{della2017controlling}.
This motivates exploration of alternative methods of certifying performance for soft robots, ones that do not necessitate asymptotic convergence to a trajectory. Inspired by the the above discussion, we explore formal guarantees in operation with the use of Control Barrier Functions (CBFs) to govern their behavior. 

Barrier Functions (BFs) are Lyapunov-like functions \cite{Tee2009,Wieland2007} and have their origins in the optimization literature \cite{Boyd2004}, in which case they are added in objective functions. Their primary use is to enforce constraints while doing optimization. Control Barrier Functions (CBFs) represent an extension of BFs tailored for control systems. They transform a constraint defined in terms of system states into a constraint on the control inputs. CBFs offer a state-feedback controller that is rigorously proven to be safe while remaining computationally efficient \cite{Aaron2014}. Specifically, CBFs are well-suited for constraints characterized by a relative degree of one concerning the system dynamics \cite{Aaron2014,Glotfelter2017}. The High Order CBF (HOCBF) \cite{Xiao2019} is designed to effectively handle constraints with arbitrarily high relative degrees, making it a versatile extension of the conventional CBF framework.

For soft-rigid robots that experience self-contact, CBFs provide a natural mechanism to design controllers that can gracefully regulate behavior near contact points (they have previously been used for something similar with humanoids \cite{khazoomHumanoid2022}). They also naturally encapsulate other constraints common in continuum robots, such as limits to extension. More generally for soft and interactive robots, CBFs provide a mechanism of safety and performance verification that can be used to guarantee properties without relying on asymptotically stable control of the state of the robot. In this work, we adopt the commonly-used Piecewise Constant Curvature (PCC) model for our system. To our knowledge, this is the first work applying CBFs to continuum robot models. 

\section{Preliminaries and System  Formulation}
\begin{figure}[t]
\centering
\includegraphics[width=0.4\textwidth]{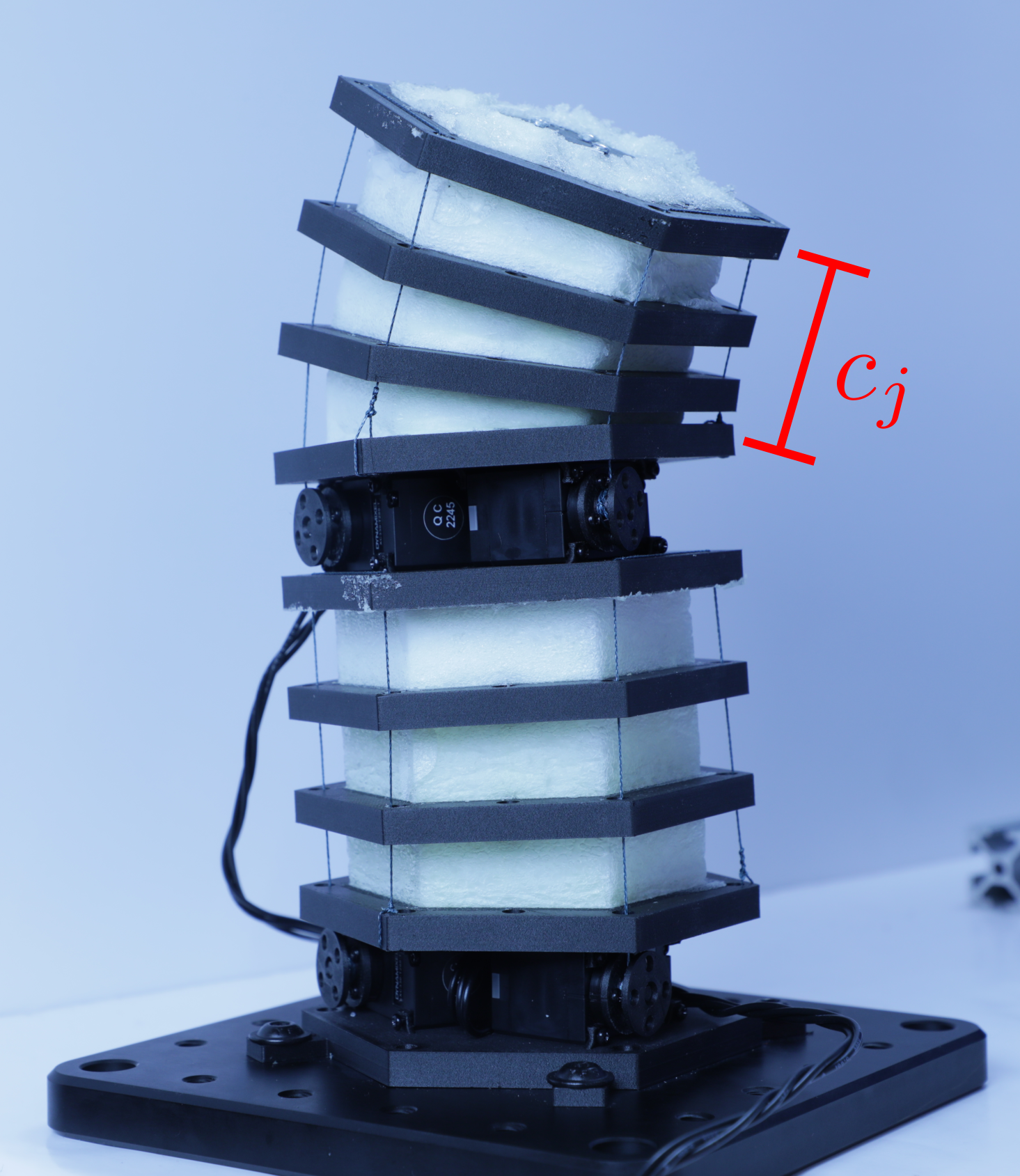}
\caption{An example of the type of hardware we examine for this work, where the modular soft-rigid segments frequently make self contact. We seek to operationalize representations of distance functions such as $c_j$ to gracefully control such structures in the presence of self contact using CBFs.}\label{fig:system}
\end{figure}

\subsection{High Order CBFs} 
\label{sec:hocbf}

We briefly introduce the background of high order CBFs \cite{Xiao2019} in this section,
and we start with some definitions for CBFs/HOCBFs. Note that this section contains formal definitions and a theorem from previous works \cite{Xiao2019,Aaron2014,Khalil2002,Glotfelter2017}, which we reproduce here for completeness.

In this paper, we consider an affine control system:
\begin{equation}
\dot{\bm{x}}=f(\bm x)+g(\bm x)\bm u \label{eqn:affine}%
\end{equation}
where $\bm x\in X\subset\mathbb{R}^{n}$, $f:\mathbb{R}^{n}\rightarrow\mathbb{R}^{n}$
and $g:\mathbb{R}^{n}\rightarrow\mathbb{R}^{n\times q}$ are
Lipschitz continuous, and $\bm u\in U\subset\mathbb{R}^{q}$ is the control constraint set.

\begin{definition}
	\label{def:forwardinv} (\cite{Aaron2014}) A set $C\subset\mathbb{R}^{n}$ is forward invariant for
	system (\ref{eqn:affine}) if solutions {for some $\bm u\in U$} starting at any $\bm x(0) \in C$
	satisfy $\bm x(t)\in C,$ $\forall t\geq0$.
\end{definition}

\begin{definition}
	\label{def:relative} (\textit{Relative degree} \cite{Khalil2002}) The relative degree of a  differentiable function $b:\mathbb{R}^{n}%
	\rightarrow\mathbb{R}$ w.r.t. system (\ref{eqn:affine}) is the number
	of times it must be differentiated along the dynamics until any component of
	$\bm u$ appears in the corresponding derivative.
\end{definition}

For constraint $b(\bm x)\geq 0$, the relative degree is the same for that of function $b(\bm x)$. 
Consider a constraint $b(\bm x)\geq0$ of relative
degree $m$, where $b:\mathbb{R}^{n}\rightarrow\mathbb{R}$, we define $\psi_{0}(\bm
x):=b(\bm x)$ and a sequence of functions $\psi_{i}:\mathbb{R}%
^{n}\rightarrow\mathbb{R},i\in\{1,\dots,m\}$ in the form:
\begin{equation}
\begin{aligned} \psi_i(\bm x) := \dot \psi_{i-1}(\bm x) + \alpha_i(\psi_{i-1}(\bm x)),i\in\{1,\dots,m\}, \end{aligned} \label{eqn:functions}%
\end{equation}
where $\alpha_{i}(\cdot),i\in\{1,\dots,m\}$ denotes a $(m-i)^{th}$ order
differentiable class $\mathcal{K}$ function \cite{Khalil2002}.

Further, we define the corresponding safe sets $C_{i}, i\in\{1,\dots,m\}$ associated
with (\ref{eqn:functions}):
\begin{equation}
\label{eqn:sets}\begin{aligned} C_i := \{\bm x \in \mathbb{R}^n: \psi_{i-1}(\bm x) \geq 0\}, i\in\{1,\dots,m\}. \end{aligned}
\end{equation}

\begin{definition}
\label{def:hocbf} (\textit{High Order Control Barrier Function (HOCBF)}
\cite{Xiao2019}) Let $C_{i}, i\in\{1,\dots, m\}$ be defined by (\ref{eqn:sets}%
) and $\psi_{i}(\bm x), i\in\{1,\dots, m\}$ be defined by
(\ref{eqn:functions}). A function $b: \mathbb{R}^{n}\rightarrow\mathbb{R}$ is
a HOCBF of relative degree $m$ for
system (\ref{eqn:affine}) if there exist $(m-i)^{th}$ order differentiable
class $\mathcal{K}$ functions $\alpha_{i},i\in\{1,\dots,m-1\}$ and a class
$\mathcal{K}$ function $\alpha_{m}$ such that 
\begin{equation}
\label{eqn:constraint}\begin{aligned} \sup_{\bm u\in U}[L_f^{m}b(\bm x) + L_gL_f^{m-1}b(\bm x)\bm u + R(b(\bm x)) \\+ \alpha_m(\psi_{m-1}(\bm x))] \geq 0, \end{aligned}
\end{equation}
for all $\bm x\in C_{1} \cap,\dots, \cap C_{m}$. The part before the inequality in
(\ref{eqn:constraint}) is actually $\psi_m(\bm x)$, $L_{f}$ ($L_{g}$) denotes Lie derivatives along
$f$ ($g$), and $R(b(\bm x)) = \sum_{i = 1}^{m-1}L_f^i(\alpha_{m-i}\circ\psi_{m-i-1})(\bm x).$ 
\end{definition}

HOCBFs generalize CBFs of relative degree one \cite{Aaron2014},
\cite{Glotfelter2017}.

\begin{theorem} [\cite{Xiao2019}]
\label{thm:hocbf}  Given a HOCBF $b(\bm x)$ from Def.
\ref{def:hocbf} with the associated sets $C_{1}, \dots, C_{m}$ defined
by (\ref{eqn:sets}), if $\bm x(0) \in C_{1} \cap,\dots,\cap C_{m}$,
then any Lipschitz continuous controller $\bm u(t)\in U$ that satisfies the constraint in
(\ref{eqn:constraint}), $\forall t\geq0$ renders $C_{1}\cap,\dots,
\cap C_{m}$ forward invariant for system (\ref{eqn:affine}).
\end{theorem}

CBFs/HOCBFs are used to transform nonlinear safety-critical control optimization problems onto a sequence of convex optimizations for system (\ref{eqn:affine})  \cite{Aaron2014}, \cite{Glotfelter2017}. We discretize the time, and hold the state as a constant within each time interval. Then, the optimization becomes a quadratic program within each time interval when the cost is quadratic in control.  The inter-sampling effect, feasibility, adaptivity, and optimality of the CBF method are extensively studied in \cite{xiao2023safe}.

With  $H\in\mathbb{R}^{q\times q}$ (positive definite) and $F\in\mathbb{R}^q$, we can include the CBF as a constraint in the QP as follows:
	\begin{equation} \label{eqn:obj}
	\begin{aligned}
	\min_{\bm u(t)\in U} \quad &\bm u^T(t) H \bm u(t) + F^T\bm u(t)\\
	\text{s.t. } \quad &\\
	L_f^{m}b(\bm x) + [L_g&L_f^{m-1}b(\bm x)]\bm u \!+\! R(b(\bm x)) + \alpha_m(\psi_{m-1}(\bm x)) \geq 0.
	\end{aligned}
	\end{equation}

The main advantage of CBFs/HOCBFs lies in their high computation efficiency for nonlinear systems \cite{Aaron2014}. They are used to guarantee system safety. In this paper, we use CBFs/HOCBFs to achieve safe manipulation for soft-rigid robots.

\subsection{Kinematics and Dynamics}

In the following work, we consider systems like those shown in Fig. \ref{fig:system}. Manipulators of this type were first presented in \cite{bernContactRich2022}. The salient features of this device are the solid foam elastic continuum and the rigid plates that make frequent self contact during operation. The segments are actuated with three motor-driven tendons arranged equally around the perimeter. For the following, in order to model this system, we adopt a Piecewise Constant Curvature approximation of the kinematics \cite{websterDesignKinematicModeling2010}. Following Della Santina \cite{dellasantinaImproved2020}, we utilize a singularity free parametrization wherein the state variables for the segment are 
\begin{equation}
    \bm q = [\Delta_\mathrm{x}, \Delta_\mathrm{y}, \delta L]^T.
\end{equation}
While these variables correspond directly to physical quantities, they are not necessarily intuitive for a human. Suffice it to say that the first two correspond to bending along the x and y axes respectively, while the third is a straightforward extension of compression of the segment. The reader is referred to \cite{dellasantinaImproved2020} for more details. In the forthcoming, we will make use of the bend angle 
\begin{equation}
    \theta = \frac{1}{d}\sqrt{\Delta_\mathrm{x}^2 + \Delta_\mathrm{y}^2},
\end{equation}
where $d$ is the radius of the segment and is chosen here to be the distance from the center of a segment to a tendon. We can easily calculate the current position of any point on the robot by using the standard forward kinematics based on our PCC model:
\begin{equation}\label{eq:fk}
    r(\bm q) = \mathrm{FK}(\bm q).
\end{equation}

While the dynamics of our system are clearly hybrid, we will assume that we do not make contact for the purposes of our dynamic model (and indeed, the purpose of our demonstration of CBFs will be to prevent the robot from driving the system through the contact point). Therefore, we can derive and write the dynamics for our PCC model in the usual form,
\begin{equation}\label{eq:dyanmics}
    M(\bm q)\bm{\ddot q} + C(q,\bm{\dot q}) \bm{\dot q} + G(\bm q) + K(\bm q) + D(\bm q)\bm{\dot q} = \bm \tau,
\end{equation}
where $M(\bm q)$ is the mass matrix, $C(q, \bm{\dot q})$ is the Coriolis matrix, $G(\bm q)$ is the gravity force, $K(\bm q)$ is the elastic force ($K(\bm q) = Kq$ when using the linear elastic assumption), $D(\bm q)$ is the damping matrix, and $\bm \tau$ is the input vector. We will now discuss the application of CBFs to this PCC soft robot model in general.

\subsection{CBFs for Soft Robots}
To reiterate, the purpose of CBFs is to enforce that a closed loop system's trajectories remain in a forward invariant set. Thus, within robotics we can use them to enforce set constraints on any property of the system that can be written as a function of the state and its derivatives (e.g. $q$, $\bm{\dot q}$, $\bm{\ddot q}$, etc), which includes most of the kinematic quantities of the robot. In this work, we will focus on safety constraints corresponding to constraints on points on the robot's body with respect to other parts of the robot's body or environment. For example, for a surgical soft robot, we might use such constraints to guarantee that the continuum does not touch a fragile piece of tissue during a minimally invasive surgery. 

To specify such a constraint, we can begin by using the forward kinematics (\ref{eq:fk}) to write down the point under consideration of our robot, $r(\bm q)$. Then, we can define our constraint with respect to the environment as,
\begin{equation}\label{eq:gen_c}
    b(\bm q) = h(r(\bm q)),
\end{equation}
where $h$ is some function, often a distance function or similar, that describes the safe set. Given this safety constraint and that this safety constraint is relative degree two \cite{Khalil2002}, we can choose both class $\mathcal{K}$ functions in (\ref{eqn:functions}) as linear functions with constant coefficient $p> 0$. The CBF constraints corresponding to (\ref{eqn:constraint}) are then
\begin{equation}\label{eqn:2CBF}
    L_f^2b + L_gL_fb \bm u+ 2p L_fb + p^2b \geq 0.
\end{equation}

All that is left is to translate the CBF into kinematic and dynamic quantities of our robot as follows. We first note that $u = \bm \tau$, $f = [\bm{\dot q}, M^{-1}(C\bm{\dot q} + G + K \bm q + D\bm{\dot q})]$ and $g = [0, M^{-1}]$ (\ref{eq:dyanmics}). Performing the Lie derivatives, we can write,
\begin{align}\label{eq:lie}
    L_fb &= \dot b = J \bm{\dot q} \\
    L_f^2b + L_gL_fb \bm u &= \ddot b = J  \bm{\ddot q} + \dot J \bm{\dot q},
\end{align}
where $J$ is the Jacobian of our constraint function (\ref{eq:gen_c}):
\begin{equation}
    J = \frac{\partial b}{\partial \bm q} = \frac{\partial h}{\partial r}\frac{\partial r}{\partial \bm q}
\end{equation}
and $\dot J$ is its time derivative.
Finally, substituting (\ref{eq:lie}) into (\ref{eqn:2CBF}), we have the following CBF constraints in terms of our PCC quantities:
\begin{equation}\label{eq:barrier}
    p^2 b + 2p J \bm{\dot q} + J M^{-1}(-C \bm{\dot q} - G - K \bm q - D \bm{\dot q} + \bm \tau) + \dot J \bm{\dot q} > 0.
\end{equation}
In the next section, we will show a specific application of this CBF for regulating self-contact of a soft-rigid hybrid robot.

\section{Control Approach}
A basic control goal for this type of system is to prevent the controller from trying to actuate through self-contact, which can potentially break the robot. In this section, we will define this safety constraint mathematically and derive a nominal controller. The control idea is naturally encompassed by Control Barrier Functions, upon which we will elaborate in the following. 

\subsection{Nominal Control Input}
For our nominal control input, we utilize a PD+ controller \cite{dellasantinaModelBasedControlSoft2023} of the following form,
\begin{multline}\label{eq:nom}
    \bm \tau_\mathrm{nom} = M(\bm q)\bm{\ddot{\overline q}} + C(\bm q,\bm{\dot q}) \bm{\dot q} + G(\bm q) + K \bm{\overline{q}} + D\bm{\dot{\overline q}}\\ + K_\mathrm{P} (\bm{\overline{q}} - \bm{q}) + K_\mathrm{D} (\bm{\dot{\overline{q}}} - \bm{\dot q}),
\end{multline}
where $\overline q$ is a desired trajectory. As demonstrated in \cite{dellasantinaModelbasedDynamicFeedback2020a}, this controller is asymptotically stable for trajectory $\overline q$ if $K_\mathrm{P}, K_\mathrm{D} > 0$. An issue with this controller for our system is that, if it allowed to operate without constraint, it can damage the physical robot by attempting to push through contact points. This inspires the use of Control Barrier Functions in the following section to attenuate the controller near self-contact.

\subsection{CBFs for Safe Self-contact}
\begin{figure}[t]
\centering
\includegraphics[width=0.3\textwidth]{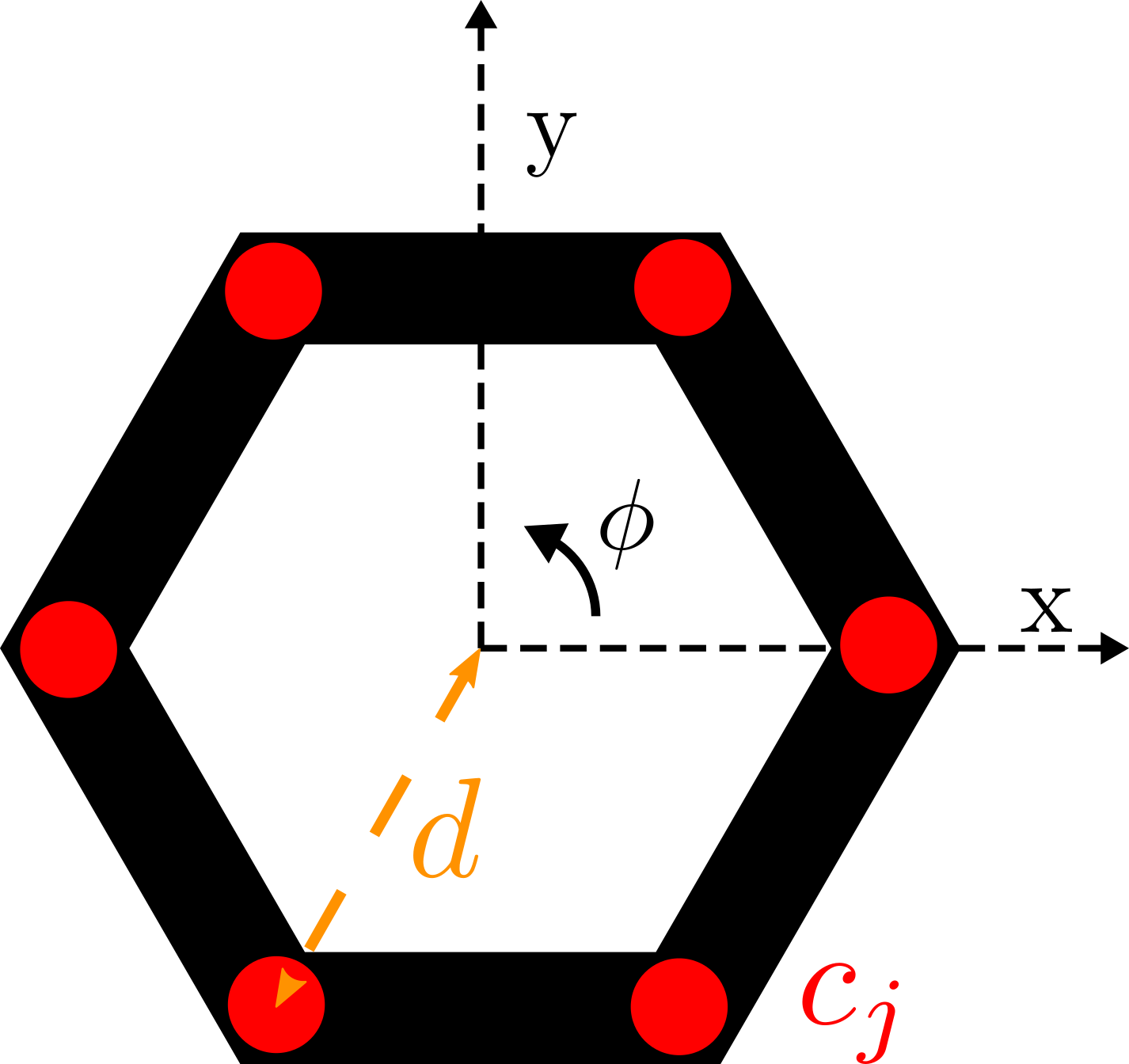}
\caption{Illustration of some key quantities for the robot. The six red dots corresponding to the corners of the plate are used as our CBFs by deriving the forward kinematics of each point. The quantity $d$ represents the distance to these corners, $\phi$ represents the angle of rotation for the vector pointing to each corner from the center.}\label{fig:hex}
\end{figure}

To specify our CBFs, for each segment (in Fig. \ref{fig:system} there are two), we take the points of the top hexagonal plates of the segment using the PCC forward kinematics from Eq. (\ref{eq:fk}). This gives the following function for each point:
\begin{multline}
    c_j(\bm q) = \frac{1}{\theta d}(\sin(\theta)(L_{0,i} d + d \delta L_i\\ - r \Delta_{\mathrm y,i} \sin(\phi_j) - r \Delta_{\mathrm x,i} \cos(\phi_j))),
\end{multline}
where $L_{0,i}$ is the uncompressed length of a segment, $\phi_j$ is the angle of the corner of the plate relative to the x-axis (shown in Fig. \ref{fig:hex}), $d$ is the distance from the center of the segment to the cable routes, and $r$ is the distance from a corner of the hexagon plate to the center of the segment. For these last two quantities, in our case, $d = r$. See Fig. \ref{fig:hex} for an illustration of some of these. The quantity $c_j$ is useful for our purposes because when $c_j > \epsilon$, where $\epsilon$ is some positive constant (a good choice is corresponding to the thickness of the plates), the plates are not in contact and they make contact as $c_j - \epsilon \rightarrow 0$.

Our barrier functions will then take the form,
\begin{equation}\label{eq:b}
    b_j(\bm q) = c_j(\bm q) - \epsilon_j,
\end{equation}
where $\epsilon_j > 0$ is a parameter to specify the distance from contact that the safety constraint should be enforced. To calculate our CBF (\ref{eq:barrier}), we also need to calculate the Jacobian of (\ref{eq:b}), which in this case is simply
\begin{equation}
    J_j(\bm q) = \frac{\partial c_j(\bm q)}{\partial \bm q}.
\end{equation}
We omit inclusion of the explicit solution of the Jacobians due to their length, but they are easily calculated using a symbolic computing package. The time derivatives of the Jacobians, $\dot J_j$, are similarly calculated. We can then form matrices by stacking the safety constraints and their Jacobians, $b(\bm q) = [b_1, ..., b_N]^T$, $J(\bm q) = [J_1, ..., J_N]^T$, $\dot J(\bm q) = [\dot J_1, ..., \dot J_N]^T$, where $N$ is the number of constraints. Note that $b$, $J$, and $\dot J$ all have limits that are well defined as $\theta \rightarrow 0$ (straight configuration). Along with the dynamic quantities from (\ref{eq:dyanmics}), we now have everything we need to calculate our CBF according to (\ref{eq:barrier}). All that is left is to encapsulate our nominal controller and CBF into a quadratic program as follows:
\begin{equation}\label{eq:qp}
\begin{aligned}
\min_{\bm \tau} \quad & \frac{1}{2}||\bm \tau - \bm \tau_\mathrm{nom}||^2\\
\textrm{s.t.} \quad & \textrm{Safety (HOCBF) Constraint (\ref{eq:barrier}).}    \\
\end{aligned}
\end{equation}

\section{Simulation Results}
\begin{figure}[t]
\centering
\includegraphics[width=0.45\textwidth]{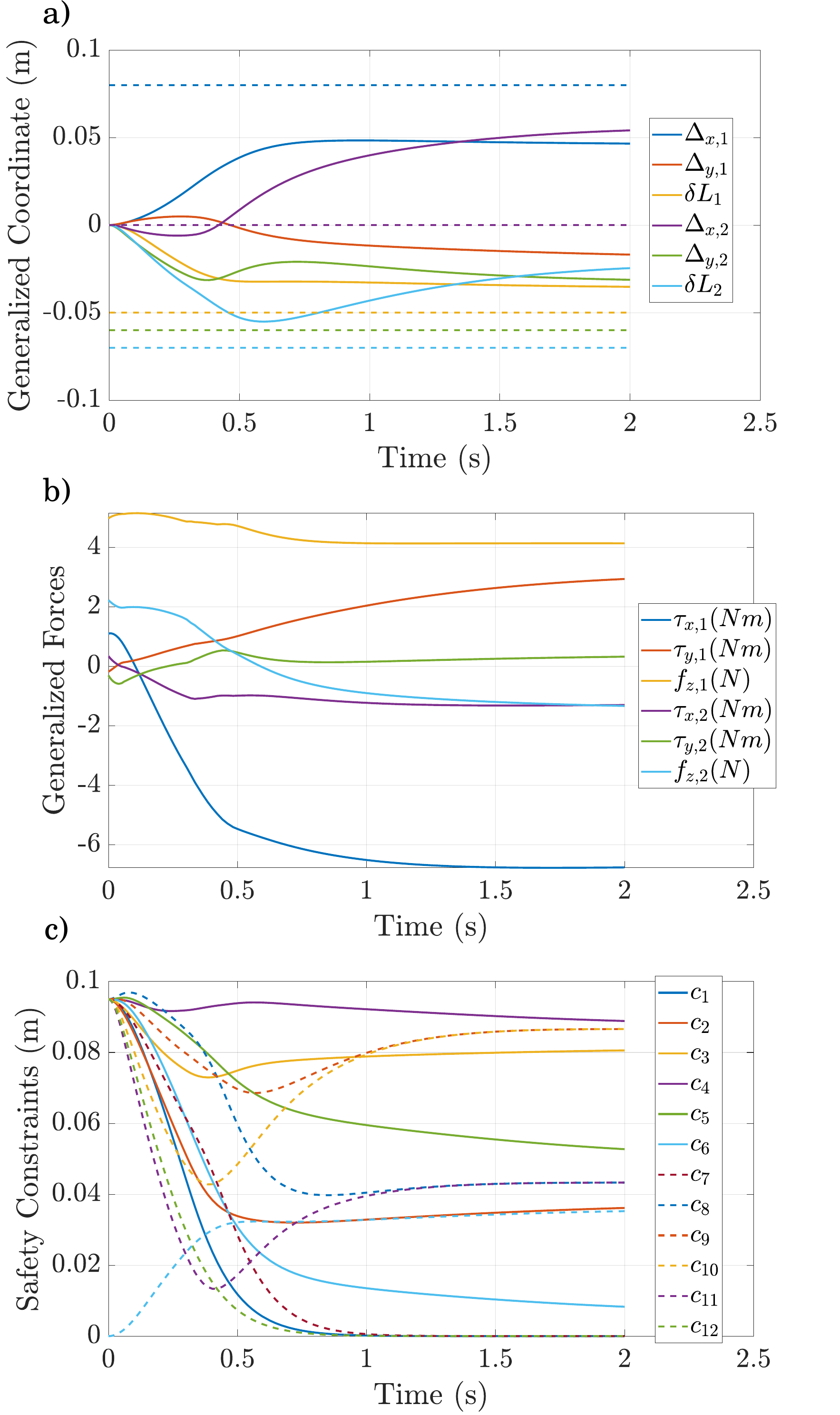}
\caption{Results in Simulation. a) Shows the generalized coordinates (solid lines) and set points (dashed lines) for a two segment simulated soft-rigid manipulator. b) Output torques from QP (\ref{eq:qp}). c) Safety constraint values during the simulation. Note that there are 6 safety constraint functions (\ref{eq:b}) per segment for a total of 12 (lines corresponding to the first and second segment are solid and dashed respectively).}\label{fig:sim}
\end{figure}
We implement the previously discussed safety constraints in a simulation to control the dynamics (\ref{eq:dyanmics}) for a two link soft-rigid hybrid manipulator. The simulator is written in Julia and we compute the mass and coriolis matrices using Featherstone's algorithms \cite{featherstone2014rigid}. Elasticity and damping are taken to be linear. We forward integrate the closed loop system using the DifferentialEquations package \cite{rackauckas2017differentialequations}, and we solve QP (\ref{eq:qp}) using the Convex package \cite{convexjl}. For both packages, the default solvers proved to be adequate for our needs. 

Parameters are set to $L_0 = 0.1 \mathrm{m}$, $d = 0.04 \mathrm{m}$, $r = 0.05 \mathrm{m}$, bending stiffness $\kappa_\theta = 10 \mathrm{\frac{Nm}{rad}}$, axial stiffness $\kappa_L = 10 \mathrm{\frac{N}{m}}$, bending damping $\beta_\theta = 5 \mathrm{\frac{Nms}{rad}}$, axial damping $\beta_L = 5 \mathrm{\frac{Ns}{m}}$, and module mass $m_j = 0.15 \mathrm{kg}$. PD gains are set to $K_\mathrm{P} = 5$ and $K_\mathrm{D} = 1$ and $\epsilon_j = 0.005 \mathrm{m}$ for all barrier functions. We simulate from initial conditions $(\bm q,\bm{\dot q}) = (\mathbf{0},\mathbf{0})$ and attempt to reach a set point $\bm{\overline q} = [0.08, 0.0, -0.05, 0.0, -0.06, -0.07]^T$. 

Results are shown in Fig. \ref{fig:sim}. We note that, as is evident from Fig. \ref{fig:sim}a, the desired set point is not possible without moving through the contact. This can be seen by observing Fig. \ref{fig:sim}c, where some safety constraint functions reach zero, but are kept from crossing the boundary due to the invariant property of CBFs. Because the CBF prevents the safety constraints from dropping below zero, it necessarily prevents states from converging to set points when doing so would counteract the safety constraint \cite{Aaron2014}, appearing in Fig. \ref{fig:sim}a as high errors for the regulator. 

\section{Hardware Results}
\begin{figure*}[t]
\centering
\includegraphics[width=0.95\textwidth]{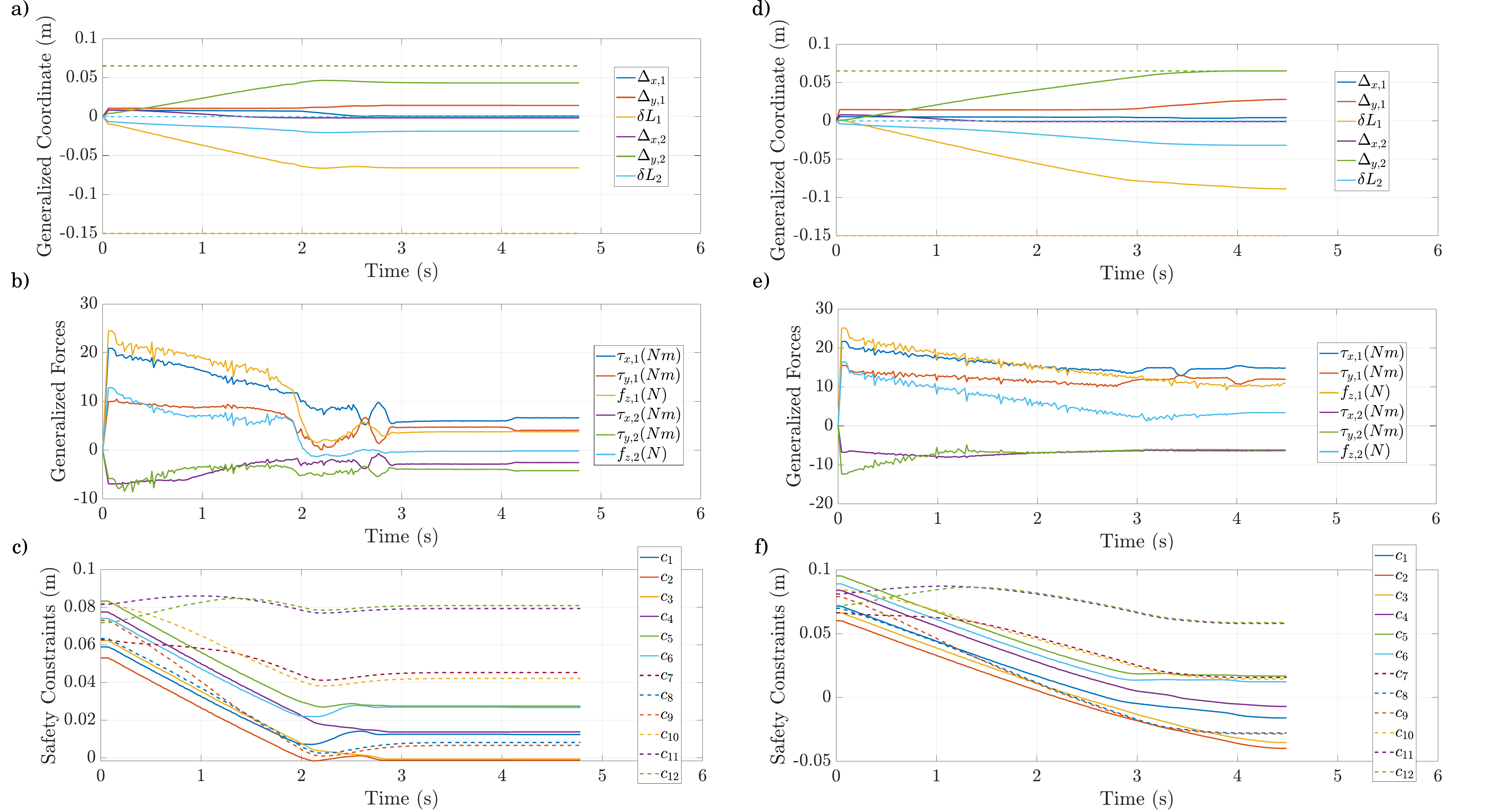}
\caption{Plots for experiments on hardware. a) Generalized coordinates and set points for two segment manipulator controlled with a CBF. b) Commanded generalized forces from Eq. (\ref{eq:qp}). c) Safety constraint values during an experiment. Note that there are 6 functions per segment for a total of 12. Note that the barrier functions are prevented from dropping below zero. d) Generalized coordinates and set points for two segment manipulator controlled with PD+. e) Commanded generalized forces from Eq. (\ref{eq:nom}). f) Safety constraints (which are not actually enforced by the nominal controller and thus are allowed to be violated).}\label{fig:hardware}
\end{figure*}
\begin{figure*}[t]
\centering
\includegraphics[width=0.85\textwidth]{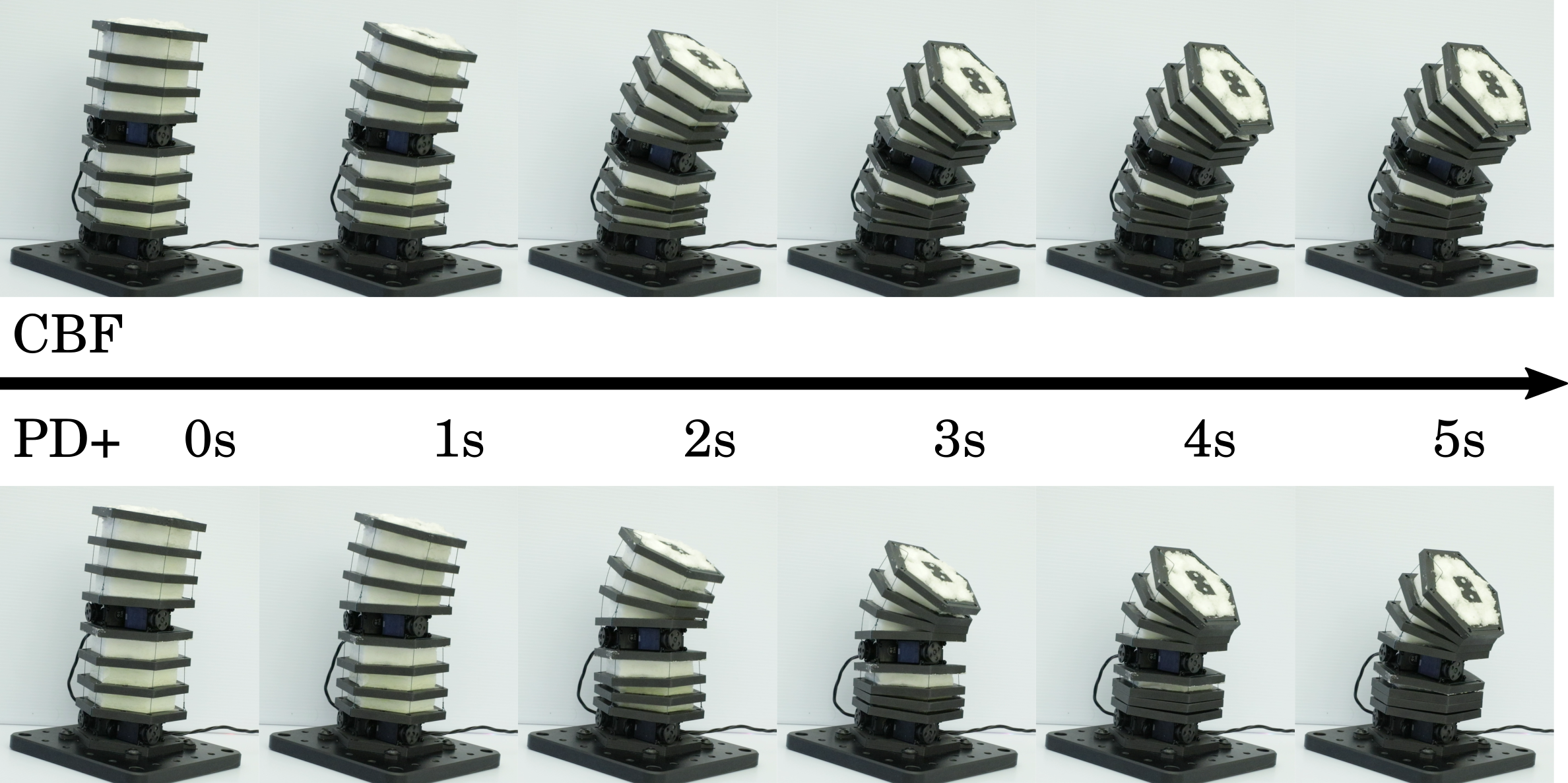}
\caption{Top: snapshots from trials controlled by QP (\ref{eq:qp}). Bottom: snapshots from PD+ controlled trial}\label{fig:snaps}
\end{figure*}
\begin{figure}[t]
\centering
\includegraphics[width=0.45\textwidth]{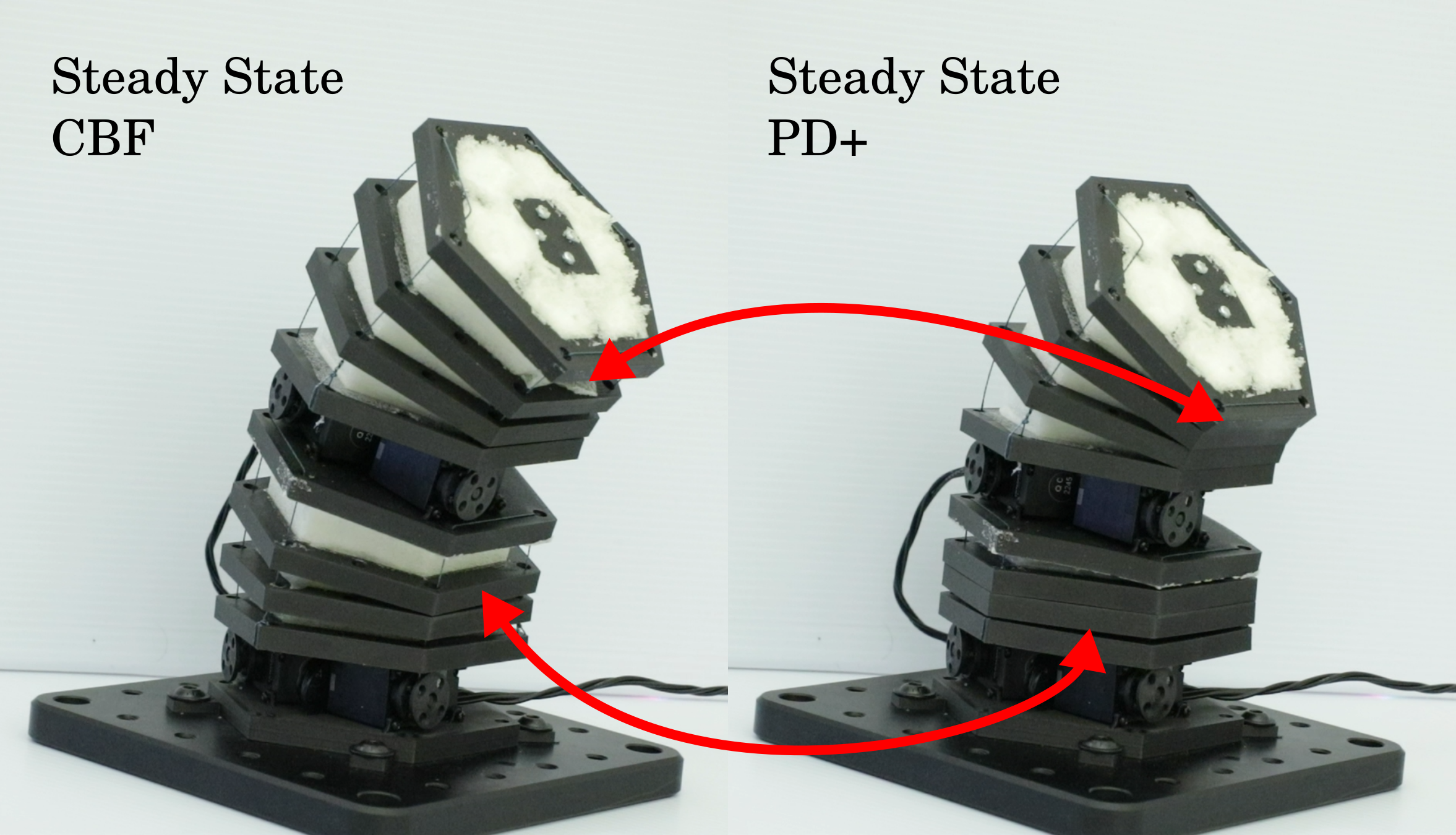}
\caption{Zoomed in images from the end of the two experiments of Fig. \ref{fig:snaps} to highlight the physical difference in convergence.}\label{fig:end_snaps}
\end{figure}

We use the hardware shown in Fig. \ref{fig:system} to validate our approach. The hardware consists of Dynamixel servo motors that actuate cables to bend and compress the PCC segments. We interface with the Dynamixels via the serial port in a Python script. To speed up the code, we calculate the mass matrix, Coriolis matrix, Jacobians, and Jacobian time derivatives using a C program that is called from Python. The QP is solved using the Python CVXPY module \cite{diamond2016cvxpy, agrawal2018rewriting}. Finally, the decision variables output by our QP are generalized forces acting on our state variables and need to be transformed such that they take the form of cable tensions. We do this using the transformation from \cite{dellasantinaImproved2020}, along with another simple transformation to account for the fact that we have three cables.

To identify the parameters of our hardware, we measure those that are readily measurable (mass, geometric properties), measure stiffness using simple feedforward experiments, and use a least squares method to identify damping and (linear) actuator gain parameters as in \cite{dellasantinaModelbased2020,ljung1998system}. For our barrier functions, $\epsilon_j = 0.005 \mathrm{m}$ for all safety constraints.  

Our code runs at $33 \mathrm{Hz}$, with the main bottleneck being the motor communication (see e.g. \cite{bestmann2019high}). We begin at rest and attempt to reach a set point $\bm{\overline q} = [0.0, 0.0, -0.15, 0.0, -0.065, 0.0]^T$, which significantly violates the specified barrier for both modules. Results are shown in Fig. \ref{fig:hardware}. The plots on the the left, Fig. \ref{fig:hardware}a-c, depict a single trial for which the CBFs are active, while those on the right, Fig. \ref{fig:hardware}d-e, depict a trial with the CBFs inactive. First, in Fig. \ref{fig:hardware}a, note that the set points for $\delta L_1$ and $\Delta_{y,2}$ are not achieved by the controller. To see why, we can observe the safety constraints in \ref{fig:hardware}c are prevented from dropping below zero, implying that the CBFs were engaged. In Fig. \ref{fig:hardware}d, we can see that $\Delta_{y,2}$ is able to reach the set point while $\delta L_1$ reaches the limit of the platform. For both modules, Fig. \ref{fig:hardware}c shows that the safety constraints are violated (drop below zero). Application of the CBF also results in a roughly 50\% difference in actuator output at steady state, as can be observed from comparing Fig. \ref{fig:hardware}b and e. We show snapshots of each trial in Fig. \ref{fig:snaps}. The last snapshots from each are enlarged in Fig. \ref{fig:end_snaps}, and it is visually evident from the gaps between the plates, or lack thereof, that the case using our CBF method is able to attenuate the controller before the contact, whereas the PD+ controller obviously does not have this capability.

\section{Discussion and Conclusion}

To the best of our knowledge, we have demonstrated the first use of Control Barrier Functions for an application in soft robotics modeled by the Piecewise Constant Curvature kinematic assumption. Specifically, we have shown that incorporating a CBF into a strategy for controlling serial soft-rigid hybrid robots is an effective strategy for regulating self contact. We derived a general CBF for a broad class of constraints on the soft robot - namely, those constraints that can be written as functions of some kinematic quantity of the robot. For our specific case of regulating self-contact, we used the CBFs to attenuate a nominal PD-style controller and implemented the solution in simulation and on hardware. Both our general approach and the specific result can easily be applied to any nominal controller, for example task space controllers \cite{khatibUnified1987,dellasantinaModelbased2020}. In future work, we will explore this approach for a broader array of systems, safety constraints, and applications.

A potential issue with this control approach is that it essentially sacrifices the nice stability property of the nominal controller when the CBF is active. The common way to deal with this is to incorporate a Control Lyapunov Function (CLF) as well, but these competing constraints may be in conflict, resulting in an infeasible QP. \cite{kurtzControl2021} provides an interesting solution to this problem using a passivity constraint, which would be interesting to implement in future work.

Finally, an open question is whether CBFs can regulate robot-environment interactions. After all, the purported point of soft robots is to allow them to make frequent, inherently safe contact with the environment. It remains to be seen whether CBFs can be used for this purpose for soft robots and this will be a topic of future work.

In conclusion, we demonstrated a Control Barrier Function workflow for regulating self contact for serial soft-rigid hybrid systems. We derived barrier functions of second degree for a Piecewise Constant Curvature system and used them in a QP based framework for effective control both in simulation and on hardware.

\section*{Acknowledgment}
 This work was done with the support of National Science Foundation EFRI program under grant number 1830901 and the Gwangju Institute of Science and Technology.

\bibliographystyle{ieeetran}
\bibliography{bib}

\begin{thebibliography}{10}
\providecommand{\url}[1]{#1}
\csname url@samestyle\endcsname
\providecommand{\newblock}{\relax}
\providecommand{\bibinfo}[2]{#2}
\providecommand{\BIBentrySTDinterwordspacing}{\spaceskip=0pt\relax}
\providecommand{\BIBentryALTinterwordstretchfactor}{4}
\providecommand{\BIBentryALTinterwordspacing}{\spaceskip=\fontdimen2\font plus
\BIBentryALTinterwordstretchfactor\fontdimen3\font minus \fontdimen4\font\relax}
\providecommand{\BIBforeignlanguage}[2]{{%
\expandafter\ifx\csname l@#1\endcsname\relax
\typeout{** WARNING: IEEEtran.bst: No hyphenation pattern has been}%
\typeout{** loaded for the language `#1'. Using the pattern for}%
\typeout{** the default language instead.}%
\else
\language=\csname l@#1\endcsname
\fi
#2}}
\providecommand{\BIBdecl}{\relax}
\BIBdecl

\bibitem{bernSimulation2022}
J.~M. Bern, F.~Zargarbashi, A.~Zhang, J.~Hughes, and D.~Rus, ``Simulation and {{Fabrication}} of {{Soft Robots}} with {{Embedded Skeletons}},'' in \emph{2022 {{International Conference}} on {{Robotics}} and {{Automation}} ({{ICRA}})}, May 2022, pp. 5205--5211.

\bibitem{zhuSoftRigid2023}
W.~Zhu, C.~Lu, Q.~Zheng, Z.~Fang, H.~Che, K.~Tang, M.~Zhu, S.~Liu, and Z.~Wang, ``A {{Soft-Rigid Hybrid Gripper With Lateral Compliance}} and {{Dexterous In-Hand Manipulation}},'' \emph{IEEE/ASME Transactions on Mechatronics}, vol.~28, no.~1, pp. 104--115, Feb. 2023.

\bibitem{coevoetPlanning2022}
E.~Coevoet, Y.~Adagolodjo, M.~Lin, C.~Duriez, and F.~Ficuciello, ``Planning of {{Soft-Rigid Hybrid Arms}} in {{Contact With Compliant Environment}}: {{Application}} to the {{Transrectal Biopsy}} of the {{Prostate}},'' \emph{IEEE Robotics and Automation Letters}, vol.~7, no.~2, pp. 4853--4860, Apr. 2022.

\bibitem{zhangGeometric2020}
J.~Zhang, T.~Wang, J.~Wang, M.~Y. Wang, B.~Li, J.~X. Zhang, and J.~Hong, ``Geometric {{Confined Pneumatic Soft}}\textendash{{Rigid Hybrid Actuators}},'' \emph{Soft Robotics}, vol.~7, no.~5, pp. 574--582, Oct. 2020.

\bibitem{dellasantinaModelBasedControlSoft2023}
C.~Della~Santina, C.~Duriez, and D.~Rus, ``Model-{{Based Control}} of {{Soft Robots}}: {{A Survey}} of the {{State}} of the {{Art}} and {{Open Challenges}},'' \emph{IEEE Control Systems Magazine}, vol.~43, no.~3, pp. 30--65, Jun. 2023.

\bibitem{pattersonRobust2022}
Z.~J. Patterson, A.~P. Sabelhaus, and C.~Majidi, ``Robust {{Control}} of a {{Multi-Axis Shape Memory Alloy-Driven Soft Manipulator}},'' \emph{IEEE Robotics and Automation Letters}, vol.~7, no.~2, pp. 2210--2217, Apr. 2022.

\bibitem{dellasantinaModelbased2020}
C.~Della~Santina, R.~K. Katzschmann, A.~Bicchi, and D.~Rus, ``Model-based dynamic feedback control of a planar soft robot: Trajectory tracking and interaction with the environment,'' \emph{The International Journal of Robotics Research}, vol.~39, no.~4, pp. 490--513, Mar. 2020.

\bibitem{dellasantinaControlling2017}
C.~Della~Santina, M.~Bianchi, G.~Grioli, F.~Angelini, M.~Catalano, M.~Garabini, and A.~Bicchi, ``Controlling {{Soft Robots}}: {{Balancing Feedback}} and {{Feedforward Elements}},'' \emph{IEEE Robotics \& Automation Magazine}, vol.~24, no.~3, pp. 75--83, Sep. 2017.

\bibitem{della2017controlling}
------, ``Controlling soft robots: balancing feedback and feedforward elements,'' \emph{IEEE Robotics \& Automation Magazine}, vol.~24, no.~3, pp. 75--83, 2017.

\bibitem{Tee2009}
K.~P. Tee, S.~S. Ge, and E.~H. Tay, ``Barrier lyapunov functions for the control of output-constrained nonlinear systems,'' \emph{Automatica}, vol.~45, no.~4, pp. 918--927, 2009.

\bibitem{Wieland2007}
P.~Wieland and F.~Allgower, ``Constructive safety using control barrier functions,'' in \emph{Proc. of 7th IFAC Symposium on Nonlinear Control System}, 2007.

\bibitem{Boyd2004}
S.~P. Boyd and L.~Vandenberghe, \emph{Convex optimization}.\hskip 1em plus 0.5em minus 0.4em\relax New York: Cambridge university press, 2004.

\bibitem{Aaron2014}
A.~D. Ames, J.~W. Grizzle, and P.~Tabuada, ``Control barrier function based quadratic programs with application to adaptive cruise control,'' in \emph{Proc. of 53rd IEEE Conference on Decision and Control}, 2014, pp. 6271--6278.

\bibitem{Glotfelter2017}
P.~Glotfelter, J.~Cortes, and M.~Egerstedt, ``Nonsmooth barrier functions with applications to multi-robot systems,'' \emph{IEEE control systems letters}, vol.~1, no.~2, pp. 310--315, 2017.

\bibitem{Xiao2019}
W.~Xiao and C.~Belta, ``Control barrier functions for systems with high relative degree,'' in \emph{Proc. of 58th IEEE Conference on Decision and Control}, Nice, France, 2019, pp. 474--479.

\bibitem{khazoomHumanoid2022}
C.~Khazoom, D.~{Gonzalez-Diaz}, Y.~Ding, and S.~Kim, ``Humanoid {{Self-Collision Avoidance Using Whole-Body Control}} with {{Control Barrier Functions}},'' in \emph{2022 {{IEEE-RAS}} 21st {{International Conference}} on {{Humanoid Robots}} ({{Humanoids}})}, Nov. 2022, pp. 558--565.

\bibitem{Khalil2002}
H.~K. Khalil, \emph{Nonlinear Systems}.\hskip 1em plus 0.5em minus 0.4em\relax Prentice Hall, third edition, 2002.

\bibitem{xiao2023safe}
W.~Xiao, C.~G. Cassandras, and C.~Belta, \emph{Safe Autonomy with Control Barrier Functions: Theory and Applications}.\hskip 1em plus 0.5em minus 0.4em\relax Springer Nature, 2023.

\bibitem{bernContactRich2022}
J.~M. Bern, L.~Z. Ya{\~n}ez, E.~Sologuren, and D.~Rus, ``Contact-{{Rich Soft-Rigid Robots Inspired}} by {{Push Puppets}},'' in \emph{2022 {{IEEE}} 5th {{International Conference}} on {{Soft Robotics}} ({{RoboSoft}})}, Apr. 2022, pp. 607--613.

\bibitem{websterDesignKinematicModeling2010}
R.~J. Webster and B.~A. Jones, ``Design and {{Kinematic Modeling}} of {{Constant Curvature Continuum Robots}}: {{A Review}},'' \emph{The International Journal of Robotics Research}, vol.~29, no.~13, pp. 1661--1683, Nov. 2010.

\bibitem{dellasantinaImproved2020}
C.~Della~Santina, A.~Bicchi, and D.~Rus, ``On an {{Improved State Parametrization}} for {{Soft Robots With Piecewise Constant Curvature}} and {{Its Use}} in {{Model Based Control}},'' \emph{IEEE Robotics and Automation Letters}, vol.~5, no.~2, pp. 1001--1008, Apr. 2020.

\bibitem{dellasantinaModelbasedDynamicFeedback2020a}
C.~Della~Santina, R.~K. Katzschmann, A.~Bicchi, and D.~Rus, ``Model-based dynamic feedback control of a planar soft robot: Trajectory tracking and interaction with the environment,'' \emph{The International Journal of Robotics Research}, vol.~39, no.~4, pp. 490--513, Mar. 2020.

\bibitem{featherstone2014rigid}
R.~Featherstone, \emph{Rigid body dynamics algorithms}.\hskip 1em plus 0.5em minus 0.4em\relax Springer, 2014.

\bibitem{rackauckas2017differentialequations}
C.~Rackauckas and Q.~Nie, ``Differential{E}quations.jl--a performant and feature-rich ecosystem for solving differential equations in {J}ulia,'' \emph{Journal of Open Research Software}, vol.~5, no.~1, 2017.

\bibitem{convexjl}
M.~Udell, K.~Mohan, D.~Zeng, J.~Hong, S.~Diamond, and S.~Boyd, ``Convex optimization in {J}ulia,'' \emph{SC14 Workshop on High Performance Technical Computing in Dynamic Languages}, 2014.

\bibitem{diamond2016cvxpy}
S.~Diamond and S.~Boyd, ``{CVXPY}: {A} {P}ython-embedded modeling language for convex optimization,'' \emph{Journal of Machine Learning Research}, vol.~17, no.~83, pp. 1--5, 2016.

\bibitem{agrawal2018rewriting}
A.~Agrawal, R.~Verschueren, S.~Diamond, and S.~Boyd, ``A rewriting system for convex optimization problems,'' \emph{Journal of Control and Decision}, vol.~5, no.~1, pp. 42--60, 2018.

\bibitem{ljung1998system}
L.~Ljung, ``System identification,'' in \emph{Signal Analysis and Prediction}.\hskip 1em plus 0.5em minus 0.4em\relax {Springer}, 1998, pp. 163--173.

\bibitem{bestmann2019high}
M.~Bestmann, J.~G{\"u}ldenstein, and J.~Zhang, ``High-frequency multi bus servo and sensor communication using the dynamixel protocol,'' in \emph{RoboCup 2019: Robot World Cup XXIII 23}.\hskip 1em plus 0.5em minus 0.4em\relax Springer, 2019, pp. 16--29.

\bibitem{khatibUnified1987}
O.~Khatib, ``A unified approach for motion and force control of robot manipulators: {{The}} operational space formulation,'' \emph{IEEE Journal on Robotics and Automation}, vol.~3, no.~1, pp. 43--53, Feb. 1987.

\bibitem{kurtzControl2021}
V.~Kurtz, P.~M. Wensing, and H.~Lin, ``Control {{Barrier Functions}} for {{Singularity Avoidance}} in {{Passivity-Based Manipulator Control}},'' in \emph{2021 60th {{IEEE Conference}} on {{Decision}} and {{Control}} ({{CDC}})}, Dec. 2021, pp. 6125--6130.

\end{thebibliography}

\end{document}